\documentclass[11pt]{article}

\usepackage[preprint]{acl}

\usepackage{times}
\usepackage{latexsym}
\usepackage[T1]{fontenc}
\usepackage[utf8]{inputenc}
\usepackage{microtype}

\usepackage{graphicx}
\usepackage{booktabs}
\usepackage{array}
\usepackage{colortbl}
\usepackage{multirow}
\usepackage{xcolor}
\usepackage{tikz}
\usetikzlibrary{positioning,arrows.meta,fit,backgrounds,calc}
\usepackage{enumitem}
\usepackage{makecell}

\definecolor{upc}{HTML}{2A78D6}
\definecolor{loc}{HTML}{EB6834}
\definecolor{badc}{HTML}{C0392B}
\definecolor{inkii}{HTML}{52514E}
\definecolor{panel}{HTML}{F4F6F9}
\definecolor{rulec}{HTML}{C9CDD3}

\newcommand{\rel}[1]{\texttt{#1}}
\newcommand{\Ds}{DeepSeek}
\newcommand{\Gp}{GPT-6-Luna}
\newcommand{\Gm}{Gemini}
\newcommand{\Qw}{Qwen}
\newcommand{\Ms}{Mistral}

\title{Evaluating Temporal Reasoning in Large Language Models:\\ Event Temporal Relation Extraction with Allen's Interval Algebra,\\ Logical Consistency, and Relative Time in Narratives}

\author{
Adib Sakhawat\textsuperscript{1$*$} \quad
Fardeen Sadab\textsuperscript{1$*$} \quad
Tamjid Hasan Fahim\textsuperscript{2$*$} \quad
Multazam Mahmud\textsuperscript{1$\dagger$} \\
Tahsan Ferdous Lihan\textsuperscript{1$\dagger$} \quad
Md Isfak Iqbal Chowdhury\textsuperscript{1$\dagger$} \quad
Nafis Ahnaf Jamil\textsuperscript{1$\ddagger$} \quad
Md. Atik Shahriar\textsuperscript{1$\ddagger$} \\
Sieam Shahriare\textsuperscript{1$\ddagger$} \quad
Wasi Omar Syed\textsuperscript{1$\ddagger$} \quad
M Safwan Hasan Khan\textsuperscript{1$\ddagger$} \quad
Tabib Hassan\textsuperscript{1$\ddagger$} \\[2pt]
\textsuperscript{1}Islamic University of Technology, Gazipur, Bangladesh \\
\textsuperscript{2}Rajshahi University of Engineering \& Technology, Rajshahi, Bangladesh
}

\newcommand\blfootnote[1]{%
  \begingroup
  \renewcommand\thefootnote{}\footnote{#1}%
  \addtocounter{footnote}{-1}%
  \endgroup
}

\begin{document}
\maketitle

\blfootnote{Authors are grouped by extent of contribution, in decreasing order:
\textsuperscript{$*$} led the work, \textsuperscript{$\dagger$} made substantial
contributions, and \textsuperscript{$\ddagger$} made supporting contributions.
Within each group, contributions are equal and the order is arbitrary.}
\blfootnote{Author addresses: \texttt{adibsakhawat}, \texttt{fardeensadab}, \texttt{multazam},
\texttt{tahsanlihan}, \texttt{isfakiqbal}, \texttt{nafisahnaf}, \texttt{atikshahriar16},
\texttt{sieamshahriare}, \texttt{wasiomar}, \texttt{safwanhasan} and
\texttt{tabibhassan}, each \texttt{@iut-dhaka.edu}; Tamjid Hasan Fahim:
\texttt{tamjidhf@gmail.com}. Correspondence to Adib Sakhawat.}

\begin{abstract}
Understanding a narrative means knowing how every event relates in time to every other, yet large language models (LLMs) are mostly tested on isolated event pairs or coarse labels. We ask five LLMs to reconstruct the complete network of Allen's 13 interval relations for 360 English narratives whose events are placed relative to one another rather than by dates, and evaluate the outputs under a pre-registered protocol for agreement with our gold annotations, converse consistency and logical satisfiability. End to end, the best model clearly beats trivial baselines on macro-F1, our primary metric, yet under our pre-registered parser no model matches the accuracy of always answering \emph{before}, which the skewed labels inflate. Most of that accuracy gap is a parsing artifact: formatting errors in large matrices cost whole rows, and a post hoc parser that recovers the determined cells of malformed rows raises coverage from 0.60 to 0.84 and closes the gap to the baseline for the best model. On the cells they do answer, models recover precedence and containment but rarely overlap, and they fail most on flashbacks, where both the narration and the event list present the later event first. Logical consistency varies widely yet says little about agreement with gold: enforcing it raises accuracy mainly by filling empty cells, and is beaten there by the trivial control of guessing the majority label. Reliable structured output and robust handling of relative time are prerequisites for temporal reasoning over whole narratives.
\end{abstract}

\begin{figure}[t]
\centering
\includegraphics[width=\columnwidth]{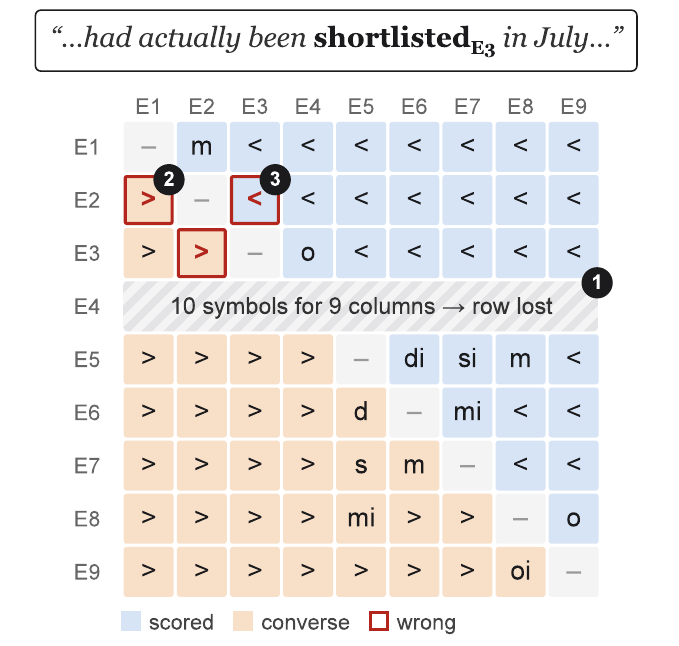}
\caption{A model must return the relation for every event pair of a passage
(schematic). Blue cells are scored against gold; orange cells must be their
converses. \textcircled{\scriptsize 1}~\textbf{Malformed rows:} one symbol too
many and the whole row is lost, so the four models with non-trivial coverage answer only 34--60\% of cells.
\textcircled{\scriptsize 2}~\textbf{Converse violations:} \textit{m} above but
$>$ below, in 0.3--37.6\% of pairs. \textcircled{\scriptsize 3}~\textbf{Consistent,
still wrong:} gold \textit{o} read as $<$/$>$; overlaps stay below 0.07~F1.}
\label{fig:overview}
\end{figure}

\section{Introduction}
\label{sec:intro}

Reading a story means building a timeline. A reader who learns, late in a passage, that a job application ``had actually been shortlisted in July'' must place that event before an interview narrated much earlier, and must keep every other event consistent with this placement. Interval relations of this kind are formalised by Allen's interval algebra \citep{allen1983maintaining}, whose 13 basic relations (\emph{before}, \emph{meets}, \emph{overlaps}, \emph{starts}, \emph{during}, \emph{finishes}, their converses, and \emph{equal}) describe every possible configuration of two intervals and come with a composition calculus for deriving new relations from known ones \citep{vilain1990constraint,nebel1995reasoning}.

Temporal relation extraction has a long history in NLP \citep{pustejovsky2003timeml,uzzaman2012tempeval,leeuwenberg2019survey}, and LLMs are now evaluated on many temporal tasks \citep{chu-etal-2024-timebench,wang-zhao-2024-tram,qiu-etal-2024-large}. Most of these evaluations, however, ask about one event pair at a time, collapse the label space to a few coarse relations, or annotate only a sparse subset of pairs \citep{cassidy-etal-2014-annotation,ning2018multi}. Pairwise querying is quadratic in the number of events and produces globally inconsistent graphs \citep{eirew2025beyond,kougia-etal-2024-analysing}, while coarse labels hide exactly the fine relations (\emph{meets}, \emph{starts}, \emph{finishes}) on which interval reasoning depends \citep{islakoglu-kalo-2025-chronosense}. Whether an LLM can reconstruct the \emph{entire} fine-grained interval network of a narrative is, to our knowledge, untested.

We study this question with a corpus of 360 English narrative passages that we annotated so that every ordered pair of events carries one of the 13 Allen relations. Five LLMs receive the passage and its event list and must return the full $n\times n$ relation matrix (Figure~\ref{fig:overview}). We fixed the prompt, the parser, the metrics and seven hypotheses in advance, and read every output in three complementary ways (Figure~\ref{fig:pipeline}, Appendix~\ref{app:setup}): the upper triangle is scored against the gold matrix, the lower triangle provides an independent converse reading of every pair, and an endpoint encoding of each network gives an exact satisfiability test and a consistent repair.

The results show that the task is far from solved and separate several reasons why. Our contributions are as follows: (i)~a pre-registered evaluation of complete Allen network extraction from narratives, in which each output supports accuracy, consistency, repair and error analyses (\S\ref{sec:protocol}); (ii)~evidence that, under single-shot free-text prompting for the whole matrix, output fidelity rather than reasoning sets the end-to-end score, since models answer only 34.5\% to 60.0\% of cells, mostly because rows have one or two symbols too many or too few, and a recovery parser restores most of the loss without new model calls (\S\ref{sec:coverage}); (iii)~evidence that consistency and correctness are decoupled, as converse violations differ by two orders of magnitude while a best-found consistent repair leaves answered-cell accuracy nearly unchanged (\S\ref{sec:consistency}); and (iv)~a diagnosis of where models fail, namely overlap relations, flashbacks that invert both list and narration order, and passages with many events, with most errors being neither one-endpoint near misses nor failures of composition (\S\ref{sec:errors}, \S\ref{sec:order}).

\section{Related Work}
\label{sec:related}

\paragraph{Temporal relations in text.}
TimeML and the TempEval campaigns established event and temporal-link annotation and evaluation \citep{pustejovsky2003timeml,uzzaman2012tempeval}, and \citet{uzzaman-allen-2011-temporal} proposed closure-based temporal awareness scores. Because annotating all pairs is quadratic, early corpora labelled few links; dense and exhaustive schemes such as TimeBank-Dense \citep{cassidy-etal-2014-annotation}, TIMELINE \citep{alsayyahi-batista-navarro-2023-timeline} and EventFull \citep{eirew-etal-2025-eventfull} cover more of each document's graph, and \citet{dsouza-ng-2014-annotating} showed that missing inter-sentence links distort evaluation. MATRES restricts judgments to start points to raise agreement \citep{ning2018multi}, and other work anchors events on a time line or on continuous scales \citep{reimers-etal-2016-temporal,vashishtha-etal-2019-fine}. Narrative corpora add causal structure \citep{mostafazadeh-etal-2016-caters,caselli-vossen-2017-event}. Our setting differs from all of these in requiring the complete set of 13 Allen relations for every event pair of a narrative.

\paragraph{Global consistency in extraction.}
Temporal relations constrain each other through symmetry and transitivity, and classical systems enforce these constraints with integer linear programming or structured learning \citep{do-etal-2012-joint,chambers-etal-2014-dense,ning-etal-2017-structured,ning-etal-2018-joint,han-etal-2019-joint,han-etal-2019-deep}. Reasoning in the full interval algebra is NP-hard, whereas point-based fragments are tractable \citep{vilain1990constraint,nebel1995reasoning}; \citet{huang-etal-2023-classification} exploit the same endpoint view to define relations as comparisons of start and end points. We use an endpoint encoding to test satisfiability exactly and to compute minimum-change repairs.

\paragraph{LLMs and temporal reasoning.}
Zero-shot ChatGPT lags supervised extractors on temporal relations \citep{yuan-etal-2023-zero}, broad benchmarks report clear gaps between LLMs and humans on ordering, duration and temporal arithmetic \citep{chu-etal-2024-timebench,wang-zhao-2024-tram}, and TNR evaluates numerical and relational temporal reasoning jointly \citep{qian-etal-2026-beyond}. \citet{qiu-etal-2024-large} find that LLMs lag behind humans and are frequently self-inconsistent, \citet{fang-etal-2024-getting} identify narrative and tense biases, and \citet{kim-hwang-2025-counterfactual} and \citet{yang-etal-2026-temporal} show that answers change when the same ordering question is rephrased. ChronoSense evaluates Allen relations between pairs of events and finds uneven performance across relations \citep{islakoglu-kalo-2025-chronosense}, QSTRBench probes composition in qualitative calculi \citep{cohn2026qstrbenchnewbenchmarkevaluate}, and TRACE frames Allen reasoning as constraint satisfaction with synthetic difficulty control \citep{zhou2026temporalreasoningbenchmarkingframework}. Closest to our work, \citet{eirew2025beyond} and \citet{zhang-etal-2024-narrative} generate whole temporal graphs with prompting, and \citet{kougia-etal-2024-analysing} show that consistent clinical predictions can still be wrong. These studies use coarse label sets, pairwise questions or synthetic inputs; we evaluate the full 13-relation network of narratives, with converse and satisfiability analyses of the same outputs.

\section{Task and Data}
\label{sec:data}

\begin{figure*}[t]
\centering
\includegraphics[width=\textwidth, trim=16 137 17 14, clip]{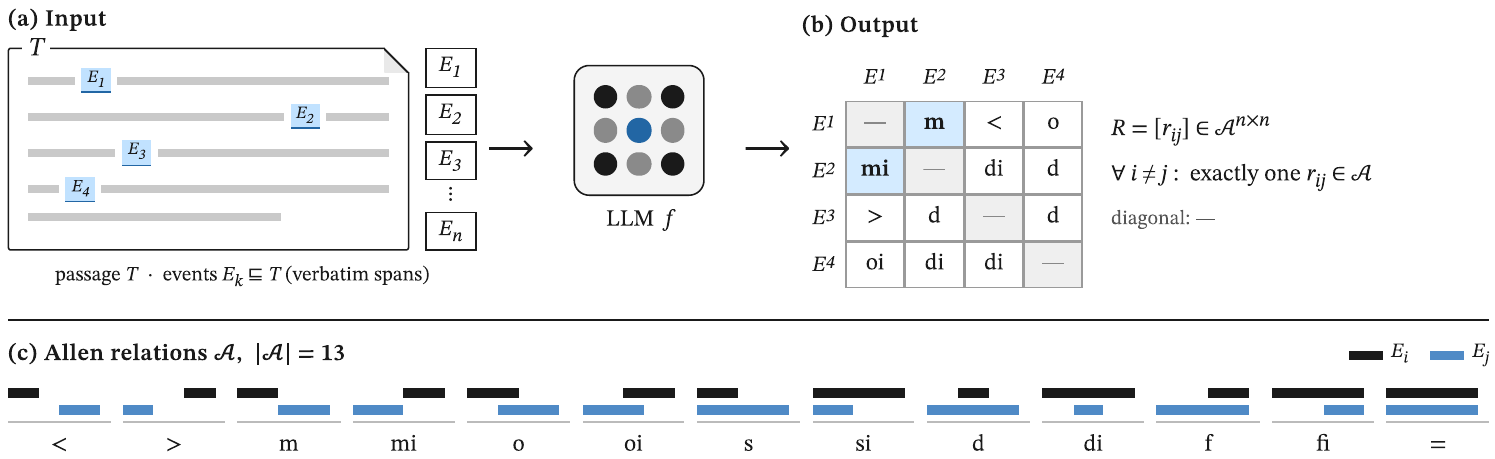}\\[-1pt]
\includegraphics[width=\textwidth, trim=16 253 17 0, clip, page=2]{figures/fig_task.pdf}
\caption{Task formulation. (a)~The input is a passage $T$ and $n$ events, each a verbatim span $E_k$ of $T$. (b)~The model returns a relation matrix $R=[r_{ij}]\in\mathcal{A}^{n\times n}$ with exactly one relation for every ordered pair and a dash on the diagonal. (c)~The 13 Allen relations between $E_i$ (black) and $E_j$ (blue). A well-formed answer is (d)~converse consistent, so that $r_{ji}=r_{ij}^{-1}$ (here $r_{12}=\rel{m}$ and $r_{21}=\rel{mi}$), and (e)~satisfiable: there are intervals $[s_k,e_k]$ with $s_k<e_k$ that realise every relation, as the timeline does for the example matrix in (b).}
\label{fig:task}
\end{figure*}

\paragraph{Task.}
Given a passage $T$ and events $E_1,\dots,E_n$, each a verbatim span of $T$, a system must assign to every ordered pair $(E_i,E_j)$, $i\neq j$, exactly one relation $r_{ij}$ from the 13 Allen relations $\mathcal{A}$ (Figure~\ref{fig:task}; Appendix~\ref{app:allen} gives definitions and the composition table). A well-formed answer is converse consistent, $r_{ji}=r_{ij}^{-1}$, and \emph{satisfiable}: some intervals $[s_k,e_k]$ with $s_k<e_k$ realise every relation.

\paragraph{Corpus.}
We wrote the English narrative passages ourselves for this study and annotated them by assigning an Allen relation to every ordered pair of selected events, so the gold matrices are complete and converse-symmetric. Most are personal narratives set in Bangladesh; the rest are non-linear literary pieces with flashbacks, procedural and engineering timelines with explicit durations, and short scenes. Time is expressed almost entirely in relative terms: only 22 passages mention a year or a day of a month (Appendix~\ref{app:relative}). We use the 360 passages with a single annotation after excluding one passage whose event list assigns the same identifier to two different events. The passages contain 130 to 1,279 words (median 241) and 8 to 78 events (median 17; 7,627 events in total), giving 101,880 unordered event pairs.

\paragraph{Properties that shape evaluation.}
Three properties of the gold data matter for interpretation. First, the label distribution is highly skewed: \rel{<} covers 60.7\% of the scored pairs (upper triangle, $i<j$ in list order) and \rel{>} 18.2\%, while \rel{oi}, \rel{s}, \rel{si} and \rel{f} each cover less than 1\% (Figure~\ref{fig:relf1} lists all supports). Predicting \rel{<} for every pair therefore yields 0.607 accuracy but 0.058 macro-F1. Second, narration order often differs from chronology: of the 80,406 gold before/after pairs, 23.9\% are narrated in reverse order; 90 passages are fully linear and 107 have more than 20\% reversed pairs. The event list follows narration order in 164 passages, and the list order and narration order of the two events agree for 93.3\% of all pairs. Third, the gold networks are redundant and nearly consistent: on average 88.7\% of a passage's pairs can be derived from two other gold relations by a single composition step, and 328 of the 360 networks are satisfiable. The 32 unsatisfiable networks contain 640 violating triples, all of the form $A\ \rel{m}\ B$, $B\ \rel{m}\ C$, $A\ \rel{m}\ C$, where the algebra requires $A\ \rel{<}\ C$. Agreement with gold remains our evaluation target in these passages, so a model that predicts the logically required relation there is counted wrong; we therefore report logical validity separately from gold agreement (\S\ref{sec:consistency}) and repeat the main results without the 32 passages (\S\ref{sec:overall}).

\section{Evaluation Protocol}
\label{sec:protocol}

Figure~\ref{fig:pipeline} in Appendix~\ref{app:setup} summarises the protocol.

\paragraph{Models.}
We evaluate five LLMs accessed through OpenRouter: DeepSeek-V4.1-Flash (\Ds), GPT-6-Luna (\Gp), Gemini-3.5-Flash-Lite (\Gm), Qwen3.7-Flash (\Qw) and Mistral-Nemo (\Ms). All models except \Ms\ accept a reasoning parameter, which we set to medium effort. We use temperature 0 where supported (\Gp\ does not accept it, so the provider default under OpenRouter's provider policy applied), seed 20261006, and an output limit of 65,536 tokens (16,384 for \Ms, its maximum). Appendix~\ref{app:setup} lists identifiers and settings.

\paragraph{Prompt and output format.}
All models receive one frozen prompt (Appendix~\ref{app:prompt}): the passage, the event list in gold identifier order, a one-line endpoint definition of each relation, and the instruction to write the complete $n\times n$ matrix between \texttt{<matrix>} tags, one line per event with $n$ space-separated symbols and \texttt{-} on the diagonal. Asking for both triangles doubles the output but gives two independent readings of every pair.

\paragraph{Pre-registration.}
Before any model was queried we froze the prompt (by hash), the gold data (by hash), the parser, the primary metric and seven hypotheses (H1 to H7, Appendix~\ref{app:hyp}), written to \texttt{PREREGISTRATION.md} before the first call and released with the code. This is a self-administered freeze without third-party timestamp. Requests were repeated only after transport failures, never because of their content. Analyses beyond H1 to H7 are exploratory, and the Holm correction applies within the pre-registered family only.

\paragraph{Parsing.}
The parser reads the last \texttt{<matrix>} block (or everything after an unclosed tag). A row is used only if it contains exactly $n$ symbols, or $n-1$ symbols when the diagonal is omitted; any other length makes the whole row unusable because its symbols cannot be aligned to columns. Unknown symbols invalidate only their cell. Unusable cells are \emph{missing}.

\paragraph{Metrics.}
We evaluate end-to-end network extraction: a relation counts only if the model wrote it in a well-formed row. The primary metric is 13-way macro-F1 over the upper triangle, with missing cells counted as wrong (a false negative for the gold label, never a false positive). Because \rel{<} covers 60.7\% of pairs, accuracy rewards majority guessing, so we report it alongside macro-F1 rather than as the headline. We also report accuracy, Cohen's $\kappa$, macro-F1 after collapsing to 7 undirected relations and to 4 classes (before, after, equal, overlap-or-contact), coverage, and accuracy and macro-F1 on answered cells only, which we treat as a conditional diagnostic rather than a measure of reasoning ability. Two trivial baselines are reported throughout: \emph{always before} and \emph{text order} (\rel{<} if the span of $E_i$ occurs first in the passage, else \rel{>}).

\paragraph{Consistency and repair.}
The converse violation rate is the share of pairs with both cells answered where $r_{ji}\neq r_{ij}^{-1}$. For satisfiability we map each relation to order constraints between the four endpoints, merge endpoints constrained to be equal, and declare the network consistent if and only if the resulting point graph has no cycle through a strict inequality; this test is exact for networks of basic relations \citep{vilain1990constraint}. Missing cells impose no constraint. The \emph{consistent repair} is the complete satisfiable network agreeing with the most answered cells that the CP-SAT solver of OR-Tools finds on the endpoint model within 30\,s per network; it is best-found and not always proven optimal. Variant (i) uses the upper triangle only; variant (ii) adds the converted lower triangle as a second vote per pair.

\paragraph{Statistics.}
Pairs within a passage are dependent, so all 95\% confidence intervals come from a cluster bootstrap over passages (1,000 resamples), and model comparisons use a paired passage-level bootstrap with Holm correction. These intervals reflect sampling of passages only; run-to-run variance is unknown (\S\ref{sec:limitations}).

\section{Results}
\label{sec:results}

\begin{table*}[t]
\centering
\small
\setlength{\tabcolsep}{4.4pt}
\begin{tabular}{l cc ccc c cc}
\toprule
& \multicolumn{5}{c}{\textbf{All 101,880 pairs} (missing = wrong)} & & \multicolumn{2}{c}{\textbf{Answered cells}} \\
\cmidrule(lr){2-6}\cmidrule(lr){8-9}
System & Accuracy [95\% CI] & Macro-F1 [95\% CI] & $\kappa$ & F1$_{7}$ & F1$_{4}$ & Coverage & Accuracy & Macro-F1 \\
\midrule
\Ds & 0.452 [0.396, 0.515] & \textbf{0.270} [0.244, 0.300] & \textbf{0.263} & \textbf{0.292} & \textbf{0.496} & \textbf{0.600} & \textbf{0.753} & \textbf{0.349} \\
\Gp & 0.350 [0.301, 0.404] & 0.215 [0.187, 0.246] & 0.183 & 0.244 & 0.391 & 0.481 & 0.728 & 0.329 \\
\Gm & 0.328 [0.284, 0.379] & 0.128 [0.113, 0.144] & 0.105 & 0.185 & 0.325 & 0.577 & 0.569 & 0.179 \\
\Qw & 0.202 [0.169, 0.242] & 0.119 [0.099, 0.141] & 0.083 & 0.157 & 0.267 & 0.345 & 0.585 & 0.232 \\
\Ms & 0.001 [0.000, 0.002] & 0.003 [0.002, 0.004] & 0.000 & 0.006 & 0.012 & 0.024 & 0.049 & 0.024 \\
\midrule
Always before & \textbf{0.607} [0.573, 0.641] & 0.058 [0.056, 0.060] & 0.000 & 0.126 & 0.189 & 1.000 & -- & -- \\
Text order & 0.601 [0.565, 0.637] & 0.072 [0.066, 0.079] & 0.053 & 0.126 & 0.235 & 1.000 & -- & -- \\
\bottomrule
\end{tabular}
\caption{Agreement with the gold matrices on the upper triangle. F1$_7$ and F1$_4$ are macro-F1 after collapsing to 7 undirected relations and to 4 classes. $\kappa$ treats a missing cell as its own category; since gold never abstains, that category appears in one margin only, depressing $\kappa$ as the missing rate grows, so these values are not comparable with $\kappa$ elsewhere. On each model's answered cells, \emph{always before} reaches 0.594 (\Ds), 0.620 (\Gp), 0.612 (\Gm), 0.624 (\Qw) and 0.633 (\Ms) accuracy, and below 0.06 macro-F1.}
\label{tab:main}
\end{table*}

\subsection{Overall agreement}
\label{sec:overall}

Table~\ref{tab:main} gives the main results. \Ds\ is the best of the five models on every metric (0.270 macro-F1, 0.452 accuracy), followed by \Gp\ (0.215). All pairwise differences in macro-F1 are significant after Holm correction ($p_{\mathrm{Holm}}=0.020$) except \Gm\ versus \Qw\ (difference 0.009, 95\% CI $[-0.009, 0.024]$, $p_{\mathrm{Holm}}=0.322$); the gap between \Ds\ and \Gp\ is 0.055 $[0.038, 0.073]$. \Ms\ produces almost no usable cells (coverage 0.024) and is excluded below.

The comparison with the baselines is the first notable result. Every model except \Ms\ exceeds both baselines on macro-F1, but no model reaches their accuracy: \Ds\ is 0.155 below \emph{always before}. A constant \rel{<} is right on 60.7\% of pairs but recovers no other relation, so models beat the baselines on macro-F1 only. Collapsing labels helps but does not change the picture: with 4 classes \Ds\ reaches 0.496 macro-F1. Excluding the 32 passages with unsatisfiable gold networks gives the same ranking with slightly higher scores (\Ds\ 0.290, \Gp\ 0.244, Appendix~\ref{app:robust}).

\subsection{Output fidelity is a first-order bottleneck}
\label{sec:coverage}

The accuracy deficit has a simple proximate cause: most models leave many cells unanswered. Coverage ranges from 0.024 (\Ms) and 0.345 (\Qw) to 0.600 (\Ds) (Table~\ref{tab:setup} gives the per-model rate of complete matrices). Truncation explains little of this \emph{overall}: 3.3\% of \Ds\ calls and 11.7\% of \Ms\ calls hit the output limit, and at most 0.8\% for the others. It is not negligible where the loss concentrates, though: 30.3\% of \Ds\ calls on passages above 40 events were truncated. The dominant failure is malformed rows. Among the matrix rows the models emitted, 18.3\% (\Ds), 29.5\% (\Gm), 33.0\% (\Gp), 42.3\% (\Qw) and 93.5\% (\Ms) have the wrong number of symbols. These are usually near misses: for \Ds, 77.5\% of malformed rows are off by exactly one symbol and 91.3\% by at most two. Counting columns reliably over long rows appears to be the limiting skill (Appendix~\ref{app:rows}).

\begin{figure}[t]
\centering
\includegraphics[width=\columnwidth]{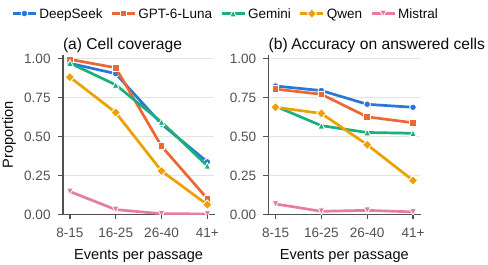}
\caption{Coverage (a) and accuracy on answered cells (b) by the number of events in the passage (152, 127, 48 and 33 passages per bin).}
\label{fig:coverage}
\end{figure}

Pairs grow quadratically with events, so the 33 passages with more than 40 events hold 39.5\% of all pairs and dominate cell-weighted coverage; averaged over passages instead, the models answer 85.4\% (\Ds) to 66.7\% (\Qw) of a typical passage. We keep the cell-weighted figure as primary, since the target is the corpus of pairs.

Coverage depends strongly on passage size (Figure~\ref{fig:coverage}a). For passages with 8 to 15 events, \Ds, \Gp\ and \Gm\ answer at least 96.8\% of cells; for passages with more than 40 events, they answer 33.6\%, 9.9\% and 31.0\%. Accuracy on answered cells also declines with size, but less steeply (Figure~\ref{fig:coverage}b): from 0.823 to 0.686 for \Ds\ and from 0.803 to 0.587 for \Gp.

\paragraph{How much of the loss is the parsing policy?}
A post hoc recovery parser (Appendix~\ref{app:recovery}) accepts the cells of a malformed row that are determined wherever the extra or missing symbol lies, then fills gaps from the model's own mirror cell. It never contradicts a strictly parsed cell, and lifts coverage from 0.600 to 0.836 (\Ds) and 0.481 to 0.870 (\Gp), with recovered cells as accurate as strict ones for \Ds\ (0.767 versus 0.753). Macro-F1 becomes 0.301 (\Ds) and 0.243 (\Gp), and \Ds\ moves from 0.155 below \emph{always before} in accuracy to parity ($+0.023$, $[-0.041, +0.086]$). Most of the end-to-end accuracy deficit is thus a property of the parsing policy, not of the models. The strict parser remains our pre-registered primary protocol.

Restricted to answered cells, the picture changes (Table~\ref{tab:main}, right). \Ds\ (0.753, 95\% CI $[0.722, 0.782]$) and \Gp\ (0.728 $[0.701, 0.754]$) clearly exceed \emph{always before} evaluated on the same cells (0.594 $[0.561, 0.629]$ and 0.620 $[0.592, 0.648]$), and their answered-cell macro-F1 (0.349 and 0.329) is about six times the baseline's. \Gm\ (0.569) and \Qw\ (0.585) remain below \emph{always before} on their own answered cells (0.612 and 0.624) even in accuracy. Answered-cell scores are conditional diagnostics: the answered cells are not a random sample, coming disproportionately from short passages, so these scores do not isolate reasoning from output fidelity and complement rather than replace the primary metric. They are also not comparable across models, since each selects its own cells. On the 22,665 cells all four answer, the ranking is unchanged and the margins widen (\Ds\ 0.805 accuracy and 0.438 macro-F1, against 0.615 and 0.059 for \emph{always before}; Appendix~\ref{app:recovery}). The ranking is not merely one by coverage, since \Gp\ exceeds \Gm\ on macro-F1 with lower coverage.

\begin{figure*}[t]
\centering
\includegraphics[width=\textwidth]{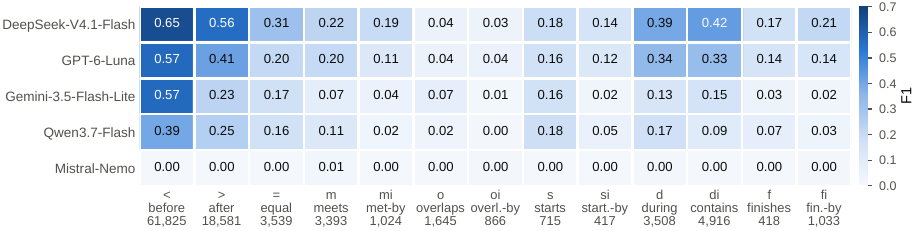}
\caption{Per-relation F1 on all pairs (missing counted as wrong). The bottom line of each column label is the number of gold pairs with that relation.}
\label{fig:relf1}
\end{figure*}

\subsection{Which relations fail}
\label{sec:relations}

Figure~\ref{fig:relf1} breaks performance down by relation. The frequent relations \rel{<} and \rel{>} and the containment pair \rel{d}/\rel{di} are recognised best (\Ds: 0.65, 0.56, 0.39 and 0.42 F1). Relations involving partial overlap are almost never recovered: \rel{o} and \rel{oi} stay at or below 0.07 F1 for every model. These estimates rest on far less support, and on few independent clusters: \rel{o} and \rel{oi} together cover 2.5\% of pairs, and \rel{oi} occurs in 67 passages against all 360 for \rel{<}, which widens the resampled macro-F1 interval. H1b predicted that \rel{s}, \rel{f}, \rel{o} and \rel{m} would be lowest; this holds for overlaps but not for \rel{s} or \rel{m}, which reach 0.16 to 0.22 F1 for the two strongest models. The confusion matrices (Appendix~\ref{app:figs}, Figure~\ref{fig:confusion}) show that answered errors flow mainly towards \rel{<}: for \Ds, 33\% of gold \rel{m} pairs and 24\% of gold \rel{o} pairs are labelled \rel{<}.

\subsection{Consistency is not correctness}
\label{sec:consistency}

\begin{table}[t]
\centering
\small
\setlength{\tabcolsep}{3.2pt}
\begin{tabular}{l ccccc}
\toprule
System & \makecell{Conv.\\viol.} & \makecell{Both\\dirs.} & \makecell{Sat.\ /\\complete} & \makecell{Triple\\viol.} & \makecell{Self-\\entail.} \\
\midrule
Gold & -- & -- & 328/360 & 0.06 & -- \\
\Ds & \textbf{0.3} & \textbf{39.9} & \textbf{212/242} & \textbf{0.12} & \textbf{0.9} \\
\Gp & 6.4 & 27.0 & 138/222 & 2.83 & 12.0 \\
\Gm & 21.3 & 25.6 & 68/207 & 3.08 & 11.0 \\
\Qw & 37.6 & 12.8 & 19/114 & 6.76 & 21.3 \\
\Ms & 90.1 & 0.0 & 1/4 & 56.26 & 34.0 \\
\bottomrule
\end{tabular}
\caption{Logical consistency (all values in \%, except satisfiable networks among complete networks). \emph{Complete} here means every cell of the \emph{upper triangle} was answered, which is a weaker requirement than the complete-matrix parse status of Table~\ref{tab:setup} (every cell of both triangles); the counts in this column are therefore higher. Conv.\ viol.: converse violations among pairs with both cells answered. Both dirs.: share of all pairs whose two cells are both correct. Triple viol.: path-inconsistent triples among checked triples. Self-entail.: answered pairs that contradict a relation entailed by the model's own answers for two other pairs.}
\label{tab:consistency}
\end{table}

Table~\ref{tab:consistency} summarises consistency, which is measured against the algebra rather than against gold. The models differ by two orders of magnitude. \Ds\ violates the converse in 0.3\% of doubly answered pairs and 87.6\% of its complete networks are satisfiable, close to the 91.1\% of the gold networks; \Qw\ violates the converse in 37.6\% of pairs and only 16.7\% of its complete networks are satisfiable. The second reading also helps \Ds: its lower triangle is more accurate than its upper triangle (0.482 versus 0.452). H2 predicted that converse violations concentrate in the non-symmetric fine relations; this holds for \Gm\ (0.511 on gold \rel{f} versus 0.147 on \rel{<}) and \Qw\ but not for \Gp, and \Ds\ has almost no violations.

Does enforcing consistency improve agreement with gold? For the four main models, the consistent repair raises accuracy on all pairs by 5.8 to 13.9 points (Figure~\ref{fig:repair}, Appendix~\ref{app:controls}), in line with H3, but the decomposition shows where the gain comes from. On the cells a model actually answered, repair changes accuracy by at most 2.5 points (\Qw, variant ii: 0.585 to 0.610) and by less than 0.1 points for \Ds\ in both variants. Nearly all of the gain comes from filling cells that were missing: the solver must assign them some consistent label, and with a skewed label distribution many of these labels are correct. Macro-F1 changes by at most 0.013 in either direction for these models; with the upper triangle alone (variant i) it decreases for \Ds, \Gp\ and \Gm. Two controls show this gain is a label-prior effect, not a logical one (Appendix~\ref{app:controls}): filling every missing cell with \rel{<} beats the solver for every model on both metrics (\Ds: 0.702 against 0.591 accuracy), and of the answered cells the solver changes, only 0.22 to 0.55 move from wrong to right. The repair objective maximises agreement with answered cells, so leaving them unchanged is partly a property of that objective. \Ds\ is nearly consistent yet wrong on 24.7\% of the cells it answers, so consistency is far from sufficient for agreement with gold, which matches the clinical findings of \citet{kougia-etal-2024-analysing}.

\subsection{What kind of errors}
\label{sec:errors}

\paragraph{Endpoint severity.}
We define the severity of an answered error as the number of the four endpoint comparisons (Table~\ref{tab:allen}) that differ from gold. H4 predicted mostly severity-1 near misses such as \rel{m} for \rel{<}; this is not the case, as severity 1 covers only 23.8\% to 33.4\% of errors and severity 2 is the most frequent class for the two strongest models. Near misses do dominate for gold \rel{o} and \rel{m} but not for \rel{>} (Appendix~\ref{app:severity}).

\paragraph{Reading or composing?}
Since 88.7\% of gold pairs follow from two other gold relations, an error on $(i,j)$ may reflect a failure to combine premises the model already had. We call an error \emph{composition-available} if some event $k$ exists for which the model labelled $(i,k)$ and $(k,j)$ correctly and their composition yields exactly the gold relation of $(i,j)$. H5 predicted that a large share of distant-pair errors are of this kind. They are rare: 4.4\% of \Ds's errors on pairs two or more sentences apart and at most 2.8\% for the other models (errors here include missing cells). The objection that correct premises are merely scarce is ruled out by a control: on cells a model gets \emph{right}, such a path is available 66.2\% to 78.1\% of the time, against 1.3\% to 4.1\% on cells it gets wrong. Premises are abundant except where the model errs, so errors come in locally corrupted neighbourhoods. Models usually lack the correct premises themselves, so the dominant problem is reading the relations out of the text, not composing them.

\subsection{Narration order and list order}
\label{sec:order}

Raw accuracy on before/after pairs is lower when the pair is narrated in reverse chronological order (\Ds: 0.415 versus 0.543; \Gp: 0.282 versus 0.436; \Gm: 0.214 versus 0.450; \Qw: 0.175 versus 0.257), as predicted by H6. On reversed pairs, 73.9\% to 76.5\% of answered errors follow narration order for \Ds, \Gp\ and \Gm, but only 47.0\% for \Qw. However, a logistic GEE model clustered by passage that controls for sentence distance and gold relation finds no significant effect of reversal for any of the four models (odds ratios between 0.65 and 2.19, all $p\geq 0.058$). That specification cannot detect the effect, however, since on before/after pairs the gold label and the list cue are one variable: pairs are ordered $i<j$, so gold \rel{>} holds exactly when the list presents the later event first, and controlling for gold removes the list cue by construction. Fitting the cue interaction instead, with sentence distance as the only control, it is significant and in the predicted direction for three of four models: odds ratio 0.222 $[0.097, 0.507]$ (\Ds), 0.096 $[0.029, 0.313]$ (\Gm), 0.200 $[0.073, 0.549]$ (\Qw), and 0.473 $[0.152, 1.473]$ (\Gp, $p{=}0.197$). The drop is specific to the conjunction of cues.

To separate the two cues, we cross list order with narration order on answered cells (Figure~\ref{fig:cue}, Appendix~\ref{app:severity}). For \Ds\ and \Gp, accuracy is stable when only one cue contradicts chronology (\Ds: 0.909, 0.923, 0.916) and drops only when both do (\Ds: 0.739 $[0.640, 0.833]$; \Gp: 0.711 $[0.635, 0.779]$). \Gm\ behaves differently: its accuracy collapses whenever the list order contradicts chronology (0.375 and 0.232), regardless of narration order, a pattern consistent with reliance on position in the event list. Flashbacks, in which the list and the narration both present the later event first, are thus the hard case for the strongest models. At the passage level macro-F1 falls from fully linear to strongly non-linear passages (\Ds: 0.393 to 0.204; Figure~\ref{fig:factors}b), but non-linearity is confounded with passage size.

\subsection{Passage size and output length}
\label{sec:length}

Larger passages are harder, and it is the number of events rather than the length of the text that predicts this. With both standardised predictors in one regression, event count is significant for all four main models ($p<0.001$) and word count is not ($p\geq 0.052$), consistent with H7: performance tracks the size of the matrix to be written, although event count is also confounded with narrative complexity. Macro-F1 for \Ds\ falls from 0.494 on passages with 8 to 15 events to 0.132 above 40 events, where 30.3\% of its calls were also truncated. Accuracy declines within an output for \Gp\ and \Qw\ but not \Ds\ (Appendix~\ref{app:severity}).

Appendix~\ref{app:extra} reports cross-model agreement, the concentration of errors on single events, and model behaviour on known gold-standard issues; notably, at least four models agree against the gold label on 665 pairs, most often reading gold \rel{m} as \rel{<}; these model--gold disagreements are candidate annotation issues, not confirmed errors.

\section{Discussion}
\label{sec:discussion}

\paragraph{Three distinct claims.}
Our results support three separate statements: end to end, models struggle to produce complete, well-formed networks; conditionally, the strongest models do much better on the cells they answer; and logical consistency does not establish agreement with gold. The end-to-end metric thus measures reasoning jointly with emitting a large, exactly formatted object, and the latter is not a formality: even the two strongest models lose 40.0\% (\Ds) and 51.9\% (\Gp) of cells to format failures under the pre-registered parser, mostly in passages with many events. Much of that loss is attributable to the protocol rather than the models: recovery parsing returns most of it (\S\ref{sec:coverage}), and asking for both triangles, which we did to obtain the converse reading, doubles the exposure to row-length errors. Scores for this task are specific to the prompting and parsing regime, and both should be stated. Since answered-cell scores alone would rank models differently (\Qw\ above \Gm), coverage, answered-cell and full scores should be reported together.

\paragraph{Consistency as a diagnostic, not a goal.}
Converse and satisfiability checks need no gold labels and separate models sharply, but are weak proxies for gold agreement: the most consistent model is still wrong on a quarter of its answers, and without per-label scores \citep{ning-etal-2017-structured,han-etal-2019-deep} a solver has little to work with.

\paragraph{Reading, not composing.}
The rarity of composition-available errors and the dominance of multi-endpoint errors suggest that failures arise when models extract relations from text, especially fine relations that hinge on shared endpoints and flashbacks that invert both surface cues. This aligns with reported narrative and position biases \citep{fang-etal-2024-getting,qiu-etal-2024-large} and suggests that interventions targeting reading, such as building a timeline before labelling \citep{zhang-etal-2024-narrative}, are more promising than symbolic post-processing.

\section{Conclusion}
\label{sec:conclusion}

Extracting complete Allen networks end to end, five LLMs reach at most 0.270 macro-F1 under our pre-registered parser, above trivial baselines on that metric but not in accuracy, mainly because big matrices are malformed; recovering the determined cells lifts the best model to 0.301 and to parity with the majority baseline in accuracy, locating most of the deficit in output format rather than reasoning. On answered cells the strongest models miss overlaps and flashbacks, and consistency says little about gold agreement: a solver enforcing it is beaten by guessing the majority label. Future work should measure run-to-run variance, test other output formats, audit model--gold disagreements and add spatial relations.

\section*{Limitations}
\label{sec:limitations}

Our confidence intervals capture variation across passages but not across repeated runs of the same model, so differences between models smaller than the unknown run-to-run variance should not be over-interpreted; \Gp\ also does not accept a temperature setting, so its provider default was used. We used one prompt, one output format (the full matrix) and medium reasoning effort, and scores, especially coverage, may differ under per-row prompting, structured-output modes, other reasoning budgets or few-shot examples, so we cannot fully separate the effect of the format from the models' reasoning ability. The five models were served through OpenRouter, which routes requests to different providers (\Ds\ calls, for example, were served by five), and because attributes such as parameter counts are not published for all models, we make no claims about scale or architecture.

The data and analyses carry further caveats. The corpus is English, mostly set in Bangladesh and limited in genre, and 101 longer passages without annotation are not evaluated. We have not yet measured inter-annotator agreement on the full corpus, so human-level agreement and the human performance ceiling remain unknown and model scores should not be read as a distance from human performance. The gold standard has known issues, and one of them is systematic rather than random. All 640 violating triples in the 32 unsatisfiable networks have the same form, $A\ \rel{m}\ B$, $B\ \rel{m}\ C$, $A\ \rel{m}\ C$, which is the pattern produced by treating \emph{meets} as transitive; the most frequent model--gold disagreement is likewise gold \rel{m} against a consensus of \rel{<} (182 of the 665 pairs). We therefore regard \rel{m} as the least reliable gold label, and not only inside the 32 passages. Because \rel{m} and \rel{mi} cover 4.3\% of the scored pairs, we repeated the main comparison with them absorbed into \rel{<} and \rel{>}: the ranking is unchanged and every model improves slightly (macro-F1 0.283, 0.227, 0.141 and 0.129 for \Ds, \Gp, \Gm\ and \Qw), as does the \emph{always before} baseline (0.071), so no conclusion in this paper depends on the \rel{m} annotations. The 665 model--gold disagreements have not been audited; model consensus alone cannot show which side is wrong. The duration flag and the temporal-expression profile in Appendix~\ref{app:relative} rely on regular expressions rather than annotation. Answered-cell scores condition on the cells a model chose to complete and are not strictly comparable across models, consistent repairs are the best found within a time limit and are not always proven optimal, and the passage-level analyses of non-linearity and size are observational and partly confounded.

\section*{Ethical Considerations}
\textbf{Data provenance and privacy.} The passages were written by the authors for this study rather than collected from people or from existing documents. They are therefore fictional: although many are personal narratives in form and are set in Bangladesh, they describe no real individuals and contain no personally identifying information, and no consent or anonymisation procedure was required. Every passage was read in full by an author during annotation, which is also our screening for offensive or objectionable content; none was found, and the genre, ordinary everyday and workplace narratives, makes such content unlikely. \textbf{Licensing.} The corpus, the model outputs, the analysis code and the paper source are released together under the MIT License. They build on no third-party data or artifacts whose terms would restrict redistribution; the evaluated models were accessed as hosted APIs and none of their weights or outputs are redistributed beyond the responses we report. \textbf{Annotation.} The temporal annotations were produced by the authors; no crowdworkers or other human participants were involved. The complete run used about USD 8.84 of API calls. Temporal relation extraction may be deployed in sensitive domains such as clinical or legal timelines \citep{kougia-etal-2024-analysing,barale-etal-2025-lextime}; our results indicate that current LLM outputs for complete interval networks are not reliable enough for unsupervised use in such settings. \textbf{Use of AI assistants.} We used AI assistants to help write code for the experiments and to improve the writing of this paper. The authors designed the study, reviewed and tested all code, checked every reported number against the stored outputs, and take full responsibility for the content.

\bibliography{custom}

\appendix

\renewcommand{\topfraction}{0.9}
\renewcommand{\bottomfraction}{0.8}
\renewcommand{\dbltopfraction}{0.9}
\renewcommand{\textfraction}{0.07}
\renewcommand{\floatpagefraction}{0.8}
\renewcommand{\dblfloatpagefraction}{0.8}
\setcounter{topnumber}{3}
\setcounter{dbltopnumber}{3}
\setcounter{totalnumber}{6}

\section{Experimental Setup and Cost}
\label{app:setup}

\begin{figure*}[t]
\centering
\includegraphics[width=\textwidth, trim=14 37 15 13, clip]{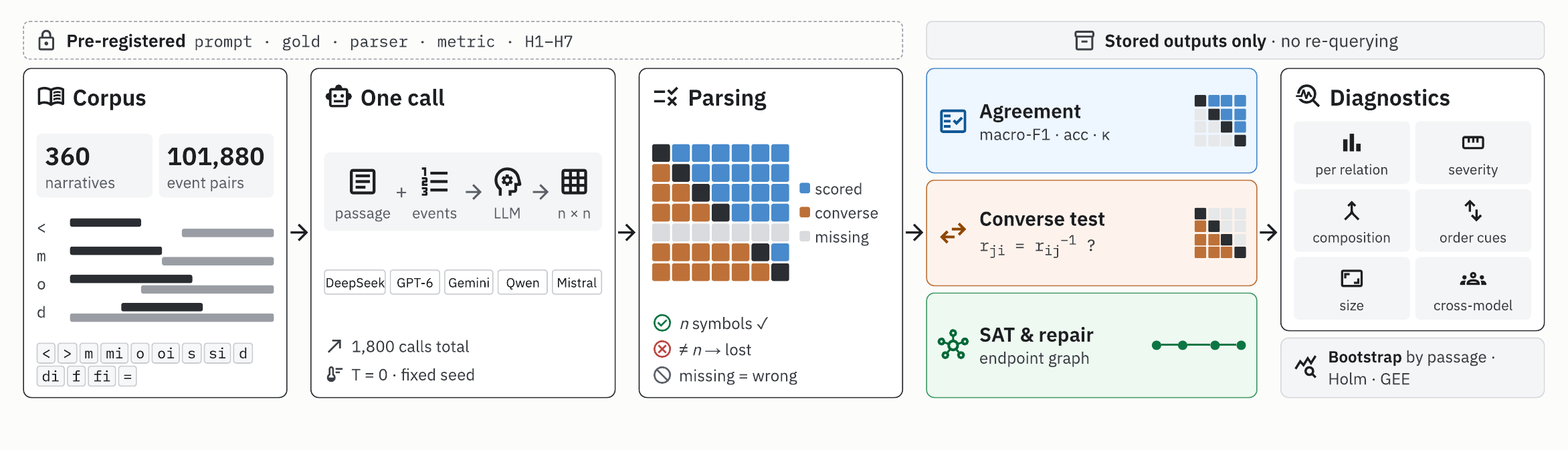}
\caption{Methodology pipeline. Everything in the dashed band was fixed in advance. Each model answers every passage, the parser turns each answer into an $n\times n$ matrix with missing cells, and all analyses read the stored matrices: the upper triangle is scored against gold, the lower triangle serves as a converse test, and the endpoint graph gives the satisfiability check and a best-found consistent repair. Uncertainty comes from a passage-level bootstrap with Holm correction and GEE models clustered by passage.}
\label{fig:pipeline}
\end{figure*}

\begin{table*}[t]
\centering
\small
\setlength{\tabcolsep}{4pt}
\begin{tabular}{l l c c c c c c}
\toprule
Model & OpenRouter identifier & \makecell{Complete /\\partial / failed (\%)} & \makecell{Trunc.\\(\%)} & \makecell{Completion\\tokens} & \makecell{Cost\\(USD)} & \makecell{Median\\latency (s)} & \makecell{Macro-F1\\per USD} \\
\midrule
\Ds & deepseek/deepseek-v4.1-flash & 66.7 / 29.7 / 3.6 & 3.3 & 7,461,014 & 4.025 & 123.2 & 0.067 \\
\Gp & openai/gpt-6-luna & 56.9 / 40.8 / 2.2 & 0.3 & 1,933,773 & 1.007 & 44.0 & 0.214 \\
\Gm & google/gemini-3.5-flash-lite & 55.3 / 42.8 / 1.9 & 0.0 & 1,188,712 & 3.076 & 11.7 & 0.042 \\
\Qw & qwen/qwen3.7-flash & 28.6 / 58.6 / 12.8 & 0.8 & 5,290,213 & 0.694 & 87.5 & 0.171 \\
\Ms & mistralai/mistral-nemo & 0.6 / 35.0 / 64.4 & 11.7 & 1,028,362 & 0.040 & 10.2 & 0.075 \\
\bottomrule
\end{tabular}
\caption{Models, reliability and cost over the 360 passages. Reasoning effort was set to medium for all models except \Ms, which has no reasoning parameter; the output limit was 65,536 tokens (16,384 for \Ms). Temperature 0 and seed 20261006 were sent to every model that accepts them (\Gp\ does not accept temperature). A call is \emph{complete} if every cell of the full $n\times n$ matrix was parsed, \emph{partial} if at least one cell is missing and \emph{failed} if no cell is usable. Table~\ref{tab:consistency} instead counts networks whose \emph{upper triangle} is complete, a weaker condition, so its counts are higher. Transport-level retries were needed for 19 (\Ds), 5 (\Gp), 20 (\Gm), 18 (\Qw) and 30 (\Ms) calls.}
\label{tab:setup}
\end{table*}

Table~\ref{tab:setup} lists the model identifiers, request settings, parse outcomes and cost. The complete run cost USD 8.84; \Gp\ offers the best macro-F1 per dollar (0.214) and \Ds\ the best absolute quality at about four times the cost of \Gp, generating 7.46M completion tokens, most of them reasoning tokens, against 1.93M for \Gp. Requests were streamed. The repair used CP-SAT with four search workers per network, a 30\,s time limit and a 3\,s stall limit; across models and variants, at most 9 of the 360 networks per configuration ended with a feasible but not proven optimal solution, and no network was left without a solution. The GEE models in \S\ref{sec:order} use a logistic link with an independence working correlation and cluster-robust standard errors. All tables and figures are generated from the stored outputs by scripts released with the paper at \url{https://github.com/sakhadib/llm-temporal-interval-reasoning}.

\section{Prompt}
\label{app:prompt}

Figure~\ref{fig:prompt} reproduces the frozen prompt. Only the passage, the event list, the number of events and the column order change between items.

\begin{figure*}[t]
\centering
\begin{minipage}{0.98\textwidth}
\footnotesize
\begin{verbatim}
===== SYSTEM =====
You are an expert annotator of temporal relations between events described in text.
===== USER =====
Read the passage and the list of events taken from it. Each event is identified by an ID and
by the words that describe it in the passage.

Your task: for every ordered pair of different events (X, Y), decide which ONE of Allen's 13
interval relations holds between the time interval of event X and the time interval of event
Y, according to the passage. Every pair must receive exactly one relation; if the passage does
not state it explicitly, choose the most plausible one.

The 13 relations (read as "X relation Y"):
<   X before Y: X ends before Y starts (there is a gap between them).
>   X after Y: X starts after Y ends (there is a gap between them).
m   X meets Y: X ends exactly when Y starts.
mi  X met by Y: X starts exactly when Y ends.
o   X overlaps Y: X starts before Y starts, and X ends after Y starts but before Y ends.
oi  X overlapped by Y: Y starts before X starts, and Y ends after X starts but before X ends.
s   X starts Y: X and Y start together, and X ends before Y ends.
si  X started by Y: X and Y start together, and X ends after Y ends.
d   X during Y: X starts after Y starts and ends before Y ends.
di  X contains Y: Y starts after X starts and ends before X ends.
f   X finishes Y: X and Y end together, and X starts after Y starts.
fi  X finished by Y: X and Y end together, and X starts before Y starts.
=   X equals Y: X and Y start together and end together.

PASSAGE:
"""
<passage text>
"""

EVENTS (<n> events):
E1: <verbatim span>
...

OUTPUT FORMAT
Write the complete <n> x <n> relation matrix between the tags <matrix> and </matrix>.
- Write exactly one line per event, in the order of the event list above.
- Each line starts with the row event's ID followed by a colon, then <n> symbols separated by
  single spaces.
- The k-th symbol in the row of event X is the relation of X to the k-th event of the list (X
  relation that event). Column order: E1 E2 ... En
- Use "-" on the diagonal (the cell of an event with itself).
- Use only these symbols: < > m mi o oi s si d di f fi = -
- Write nothing else inside the tags.

Format illustration for a different, 3-event case (the symbols here are only an illustration):
<matrix>
E1: - < m
E2: > - d
E3: mi di -
</matrix>
\end{verbatim}
\end{minipage}
\caption{The frozen prompt. Angle-bracketed placeholders stand for item-specific content; long lines of the task description are wrapped here for display.}
\label{fig:prompt}
\end{figure*}

\section{Malformed Rows}
\label{app:rows}

\begin{table}[!ht]
\centering
\small
\setlength{\tabcolsep}{3.5pt}
\begin{tabular}{l rrr cc}
\toprule
Model & Usable & Short & Long & $|\Delta|=1$ & $|\Delta|\leq 2$ \\
\midrule
\Ds & 5,774 & 172 & 1,121 & 77.5 & 91.3 \\
\Gp & 5,018 & 2,158 & 318 & 46.4 & 62.0 \\
\Gm & 5,379 & 421 & 1,825 & 66.3 & 77.2 \\
\Qw & 3,887 & 968 & 1,878 & 40.6 & 60.9 \\
\Ms & 360 & 2,791 & 2,378 & 11.5 & 20.6 \\
\bottomrule
\end{tabular}
\caption{Matrix rows emitted for the 7,627 events, by whether they have a usable length or too few (short) or too many (long) symbols. The last two columns give the percentage of malformed rows that miss the required length by one symbol or by at most two.}
\label{tab:rows}
\end{table}

Table~\ref{tab:rows} counts the rows each model emitted (first occurrence of each event identifier). Rows can also be absent, for example after truncation or when a model stops early, which is why the totals are below 7,627. \Ds\ and \Gm\ mostly write too many symbols (86.7\% and 81.3\% of their malformed rows), whereas \Gp\ mostly writes too few (87.2\%).

\section{Pre-registered Hypotheses}
\label{app:hyp}

\begin{table*}[t]
\centering
\small
\begin{tabular}{p{0.43\textwidth} p{0.50\textwidth}}
\toprule
Hypothesis (fixed in advance) & Outcome \\
\midrule
H1a: models beat the trivial baselines by a wide margin on macro-F1 but much less on accuracy. & Not supported as stated. Four models exceed the baselines on macro-F1, but under the pre-registered parser all fall \emph{below} them in accuracy; under the recovery parser the best model is level with them (\S\ref{sec:coverage}). \\
H1b: the rare relations \rel{s}, \rel{f}, \rel{o}, \rel{m} score lowest. & Partly supported: \rel{o} and \rel{oi} are lowest; \rel{s} and \rel{m} are not. \\
H2: converse violations concentrate in the non-symmetric fine relations (\rel{s}/\rel{si}, \rel{f}/\rel{fi}, \rel{d}/\rel{di}). & Partly supported: true for \Gm\ and \Qw, not for \Gp; \Ds\ has almost no violations. \\
H3: offline repair raises accuracy slightly, but consistency is not the same as correctness. & Supported, with two qualifications: the accuracy gain comes almost entirely from filling missing cells, and a control that fills them with the majority label beats the solver for every model. \\
H4: most errors are near misses differing in one endpoint comparison. & Not supported: 23.8\% to 33.4\% of errors have severity 1. \\
H5: a large share of errors on distant pairs are composition-available. & Not supported: at most 4.4\%, against 66--78\% availability on cells the models get right, so the low rate is not an artifact of premise scarcity. \\
H6: accuracy drops on pairs narrated in reverse order, and errors there mostly follow text order. & Partly supported: raw drops for all four main models and text-order errors above 50\% for three. Controlling for the gold relation removes the effect, but that control is collinear with the list-order cue; fitting the cue interaction instead, the conjunction of reversed list and reversed narration is significant for three of four models (\S\ref{sec:order}). \\
H7: F1 falls with event count, partly through output length (late rows, truncation). & Supported for event count and truncation; within-output decline for \Gp\ and \Qw\ but not \Ds. \\
\bottomrule
\end{tabular}
\caption{Pre-registered hypotheses and outcomes, based on the four models with non-trivial coverage. Holm correction applies within this family; analyses reported elsewhere in the paper that are not listed here are exploratory.}
\label{tab:hyp}
\end{table*}

Table~\ref{tab:hyp} lists the seven hypotheses and their outcomes.

\section{Robustness}
\label{app:robust}

\begin{table}[!ht]
\centering
\small
\setlength{\tabcolsep}{3.2pt}
\begin{tabular}{l ccccc}
\toprule
Model & All & \makecell{Gold-\\consistent} & \makecell{Fully\\linear} & \makecell{Non-\\linear} & \makecell{Excl.\\548, 550} \\
\midrule
\Ds & 0.270 & 0.290 & 0.393 & 0.247 & 0.273 \\
\Gp & 0.215 & 0.244 & 0.369 & 0.188 & 0.218 \\
\Gm & 0.128 & 0.143 & 0.226 & 0.110 & 0.129 \\
\Qw & 0.119 & 0.138 & 0.187 & 0.104 & 0.121 \\
\Ms & 0.003 & 0.003 & 0.003 & 0.002 & 0.003 \\
\bottomrule
\end{tabular}
\caption{Macro-F1 on subsets: all 360 passages, the 328 passages with satisfiable gold networks, the 90 fully linear and 270 non-linear passages, and all passages except two that repeat event wording.}
\label{tab:robust}
\end{table}

The model ranking is identical on every subset (Table~\ref{tab:robust}).

\section{Additional Analyses}
\label{app:extra}

\paragraph{Where errors sit.}
The end of the first event compared with the start of the second is the endpoint comparison most often wrong (68.3\% to 74.3\% of answered errors for the four main models). Errors are moderately concentrated: in passages with at least five errors, the single event with most errors accounts on average for 31.2\% (\Ds) and 29.5\% (\Gp) of the passage's errors, and for more than half of them in 12.2\% and 11.4\% of passages. Surprisingly, pairs in the same or adjacent sentences have higher error rates than distant pairs (\Ds: 0.608 versus 0.510), partly because nearby pairs contain more fine relations: macro-F1 is lowest for same-sentence pairs for every model (Figure~\ref{fig:factors}a). Events in the middle of a passage are slightly harder for all four models (\Ds: 0.433 versus 0.452 at the beginning and 0.472 at the end), a small untested pattern reminiscent of \citet{liu-etal-2024-lost}.

\paragraph{Explicit durations.}
A heuristic duration flag makes little difference to accuracy (\Ds: 0.458 with against 0.445 without).

\paragraph{Cross-model agreement.}
\label{sec:crossmodel}
Models agree with each other only moderately (Fleiss' $\kappa=0.219$ on the 1,778 pairs answered by all five models). The two strongest models agree most ($\kappa=0.652$ on cells both answered), more than either agrees with the gold standard (0.575 and 0.523, Figure~\ref{fig:kappacost}a). A plurality vote over the five models reaches 0.561 accuracy, higher than any single model, but 0.242 macro-F1, lower than \Ds. For 37.8\% of pairs no model is correct, and an oracle that picks any correct model reaches 0.622 accuracy.

\paragraph{Known gold-standard issues.}
Model behaviour on known gold problems is informative. On the 291 $(A,C)$ pairs of the $A\ \rel{m}\ B\ \rel{m}\ C$ chains whose gold label \rel{m} is logically impossible, models reproduce the gold label at most 8.6\% of the time (\Ds) and predict the logically required \emph{before} relation in 7.9\% (\Qw) to 26.1\% (\Gm) of cases. Scores on the 32 gold-inconsistent passages are much lower than on the rest (\Ds: 0.200 versus 0.290 macro-F1). We also find 665 pairs on which at least four models agree on the same label against the gold label; the most frequent pattern (182 pairs) is gold \rel{m} versus consensus \rel{<}. We call these model--gold disagreements: they are candidate annotation issues, not confirmed errors, because consensus may equally reflect a shared model bias towards \rel{<}, and their correctness cannot be determined from model consensus alone.

\section{Additional Figures}
\label{app:figs}

\begin{figure*}[t]
\centering
\includegraphics[width=\textwidth]{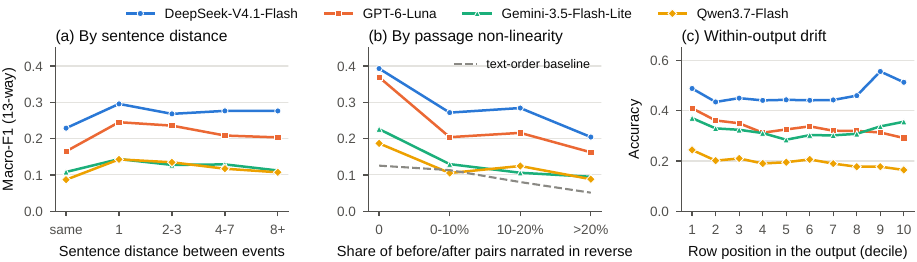}
\caption{(a) Macro-F1 by sentence distance between the two events. (b) Macro-F1 by passage non-linearity (90, 106, 57 and 107 passages), with the text-order baseline. (c) Accuracy by the position of the row in the generated matrix (missing cells counted as wrong).}
\label{fig:factors}
\end{figure*}

\begin{figure*}[t]
\centering
\includegraphics[width=\textwidth]{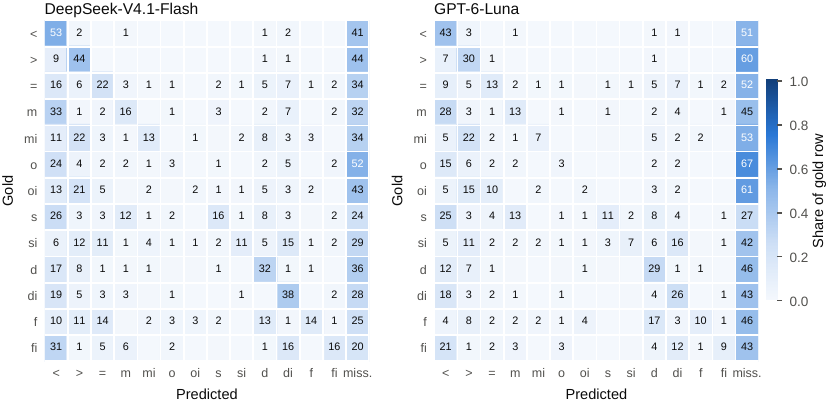}
\caption{Row-normalised confusion matrices (\%) for the two strongest models, including missing cells. Values below 0.5\% are omitted.}
\label{fig:confusion}
\end{figure*}

\begin{figure*}[t]
\centering
\includegraphics[width=\textwidth]{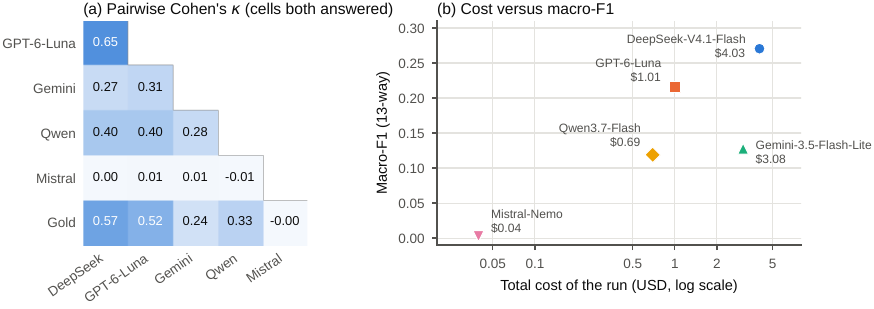}
\caption{(a) Cohen's $\kappa$ between systems, computed on the cells both systems answered (the gold annotation answers every cell). (b) Total cost against macro-F1.}
\label{fig:kappacost}
\end{figure*}

Figures~\ref{fig:factors}, \ref{fig:confusion} and \ref{fig:kappacost} complement \S\ref{sec:errors} to \S\ref{sec:length} and Appendix~\ref{app:extra}.

\section{Allen's Interval Algebra}
\label{app:allen}

\paragraph{Origins.}
\citet{allen1983maintaining} introduced the interval algebra to represent and reason about the temporal structure of events, motivated by natural language understanding and planning, where exact dates and durations are rarely given. Its primitive is the interval rather than the instant, which matches how people describe events (``during the trip'', ``as the game ended''), and its relations are purely qualitative: they carry no information about durations. Allen paired the representation with a constraint-propagation algorithm and illustrated applications to historical databases, process modelling and interactive systems. The point algebra of \citet{vilain1990constraint} offered a point-based alternative and clarified which fragments of the interval algebra remain tractable. The algebra has since become a reference calculus for qualitative temporal reasoning, for temporal information extraction \citep{leeuwenberg2019survey}, and for recent evaluations of LLMs \citep{islakoglu-kalo-2025-chronosense,cohn2026qstrbenchnewbenchmarkevaluate,zhou2026temporalreasoningbenchmarkingframework}.

\paragraph{Basic relations.}
For intervals $X=[x^-,x^+]$ and $Y=[y^-,y^+]$ with $x^-<x^+$ and $y^-<y^+$, exactly one of the 13 basic relations in Table~\ref{tab:allen} holds: the relations are jointly exhaustive and pairwise disjoint. They form six converse pairs plus the self-converse \rel{=}, and $X\ r\ Y$ holds if and only if $Y\ r^{-1}\ X$. We use the symbols of Table~\ref{tab:allen} throughout the paper and in the prompt (Figure~\ref{fig:prompt}).

\begin{table}[!ht]
\centering
\footnotesize
\setlength{\tabcolsep}{2.4pt}
\begin{tabular}{l l l cccc}
\toprule
$r$ & Name & Condition & ss & se & es & ee \\
\midrule
\rel{{<}} & before & $x^+<y^-$ & $<$ & $<$ & $<$ & $<$ \\
\rel{{>}} & after & $y^+<x^-$ & $>$ & $>$ & $>$ & $>$ \\
\rel{m} & meets & $x^+=y^-$ & $<$ & $<$ & $=$ & $<$ \\
\rel{mi} & met-by & $y^+=x^-$ & $>$ & $=$ & $>$ & $>$ \\
\rel{o} & overlaps & $x^-<y^-<x^+<y^+$ & $<$ & $<$ & $>$ & $<$ \\
\rel{oi} & overlapped-by & $y^-<x^-<y^+<x^+$ & $>$ & $<$ & $>$ & $>$ \\
\rel{s} & starts & $x^-=y^-,\ x^+<y^+$ & $=$ & $<$ & $>$ & $<$ \\
\rel{si} & started-by & $x^-=y^-,\ y^+<x^+$ & $=$ & $<$ & $>$ & $>$ \\
\rel{d} & during & $y^-<x^-,\ x^+<y^+$ & $>$ & $<$ & $>$ & $<$ \\
\rel{di} & contains & $x^-<y^-,\ y^+<x^+$ & $<$ & $<$ & $>$ & $>$ \\
\rel{f} & finishes & $x^+=y^+,\ y^-<x^-$ & $>$ & $<$ & $>$ & $=$ \\
\rel{fi} & finished-by & $x^+=y^+,\ x^-<y^-$ & $<$ & $<$ & $>$ & $=$ \\
\rel{=} & equal & $x^-=y^-,\ x^+=y^+$ & $=$ & $<$ & $>$ & $=$ \\
\bottomrule
\end{tabular}
\caption{The 13 basic relations for $X\ r\ Y$ (listed in converse pairs), their defining conditions and endpoint signatures: the order of $x^-$ versus $y^-$ (ss), $x^-$ versus $y^+$ (se), $x^+$ versus $y^-$ (es) and $x^+$ versus $y^+$ (ee).}
\label{tab:allen}
\end{table}

\paragraph{Endpoint signatures.}
Each basic relation is determined by the order of four endpoint pairs (Table~\ref{tab:allen}, right). Of the $3^4=81$ possible sign patterns only these 13 can be realised by two proper intervals. We use the signatures in three places: the satisfiability test translates every relation into constraints between endpoints (\S\ref{sec:protocol}); the consistent repair searches over endpoint orders; and the error severity in \S\ref{sec:errors} counts how many of the four comparisons differ between a predicted and a gold relation. Under this measure \rel{m} and \rel{<} differ only in es (severity 1), whereas \rel{<} and \rel{>} differ in all four (severity 4).

\paragraph{Composition.}
If $X\ r_1\ Y$ and $Y\ r_2\ Z$, the relation between $X$ and $Z$ is constrained to the set $r_1\circ r_2$ listed in Table~\ref{tab:composition}. Because intervals have two endpoints, a composition is often a set rather than a single relation: for example, $\rel{o}\circ\rel{o}=\{\rel{<},\rel{m},\rel{o}\}$, while $\rel{m}\circ\rel{m}=\{\rel{<}\}$ and $\rel{<}\circ\rel{>}$ allows all 13 relations. Of the 169 compositions of basic relations, 97 yield a single relation. These single-valued entries define the one-step derivability of gold pairs in \S\ref{sec:data} (88.7\% of pairs on average) and the composition-available errors in \S\ref{sec:errors}. We derived the table, like the endpoint signatures, by enumerating all configurations of three concrete intervals with integer endpoints, so it contains no transcription errors.

\begin{table*}[t]
\centering
\scriptsize
\setlength{\tabcolsep}{1.6pt}
\renewcommand{\arraystretch}{1.25}
\begin{tabular}{>{\rmfamily\bfseries}l *{13}{>{\ttfamily\centering\arraybackslash}p{0.064\textwidth}}}
\toprule
& \textrm{\textbf{{<}}} & \textrm{\textbf{m}} & \textrm{\textbf{o}} & \textrm{\textbf{fi}} & \textrm{\textbf{di}} & \textrm{\textbf{s}} & \textrm{\textbf{=}} & \textrm{\textbf{si}} & \textrm{\textbf{d}} & \textrm{\textbf{f}} & \textrm{\textbf{oi}} & \textrm{\textbf{mi}} & \textrm{\textbf{{>}}} \\ \midrule
\rowcolor{panel} \textbf{{<}} & {<} & {<} & {<} & {<} & {<} & {<} & {<} & {<} & {<} m o s d & {<} m o s d & {<} m o s d & {<} m o s d & \textit{full} \\
\textbf{m} & {<} & {<} & {<} & {<} & {<} & m & m & m & o s d & o s d & o s d & fi = f & di si oi mi {>} \\
\rowcolor{panel} \textbf{o} & {<} & {<} & {<} m o & {<} m o & {<} m o fi di & o & o & o fi di & o s d & o s d & \textit{conc} & di si oi & di si oi mi {>} \\
\textbf{fi} & {<} & m & o & fi & di & o & fi & di & o s d & fi = f & di si oi & di si oi & di si oi mi {>} \\
\rowcolor{panel} \textbf{di} & {<} m o fi di & o fi di & o fi di & di & di & o fi di & di & di & \textit{conc} & di si oi & di si oi & di si oi & di si oi mi {>} \\
\textbf{s} & {<} & {<} & {<} m o & {<} m o & {<} m o fi di & s & s & s = si & d & d & d f oi & mi & {>} \\
\rowcolor{panel} \textbf{=} & {<} & m & o & fi & di & s & = & si & d & f & oi & mi & {>} \\
\textbf{si} & {<} m o fi di & o fi di & o fi di & di & di & s = si & si & si & d f oi & oi & oi & mi & {>} \\
\rowcolor{panel} \textbf{d} & {<} & {<} & {<} m o s d & {<} m o s d & \textit{full} & d & d & d f oi mi {>} & d & d & d f oi mi {>} & {>} & {>} \\
\textbf{f} & {<} & m & o s d & fi = f & di si oi mi {>} & d & f & oi mi {>} & d & f & oi mi {>} & {>} & {>} \\
\rowcolor{panel} \textbf{oi} & {<} m o fi di & o fi di & \textit{conc} & di si oi & di si oi mi {>} & d f oi & oi & oi mi {>} & d f oi & oi & oi mi {>} & {>} & {>} \\
\textbf{mi} & {<} m o fi di & s = si & d f oi & mi & {>} & d f oi & mi & {>} & d f oi & mi & {>} & {>} & {>} \\
\rowcolor{panel} \textbf{{>}} & \textit{full} & d f oi mi {>} & d f oi mi {>} & {>} & {>} & d f oi mi {>} & {>} & {>} & d f oi mi {>} & {>} & {>} & {>} & {>} \\
\bottomrule
\end{tabular}
\caption{Composition table: the cell in row $r_1$ and column $r_2$ lists the possible relations between $X$ and $Z$ given $X\ r_1\ Y$ and $Y\ r_2\ Z$. \textit{full} denotes all 13 relations and \textit{conc} the nine relations in which the two intervals share time (\rel{o}, \rel{oi}, \rel{s}, \rel{si}, \rel{d}, \rel{di}, \rel{f}, \rel{fi}, \rel{=}).}
\label{tab:composition}
\end{table*}

\paragraph{Indefinite relations and complexity.}
When the relation between two intervals is only partly known, it is expressed as a disjunction of basic relations, so there are $2^{13}=8{,}192$ general relations. Deciding whether a network of general relations is satisfiable is NP-complete, and path consistency, which repeatedly intersects each relation with the compositions along all paths of length two, does not detect every inconsistency \citep{vilain1990constraint}. This motivated the search for tractable fragments: the ORD-Horn subclass is a maximal tractable subclass in which path consistency decides satisfiability \citep{nebel1995reasoning}, later work identified maximal tractable subclasses that also admit metric constraints \citep{drakengren1997eight}, and fuzzy extensions handle events with gradual or vague boundaries \citep{schockaert2008fuzzifying}. The networks in this paper are simpler: every pair carries exactly one basic relation, and basic relations translate directly into point constraints, so satisfiability reduces to cycle detection in a graph over the $2n$ endpoints and is decided exactly in polynomial time (\S\ref{sec:protocol}). Only the consistent repair, which minimises the number of changed labels, requires combinatorial search.

\section{Recovery Parsing and Comparable Subsets}
\label{app:recovery}

\paragraph{The recovery parser.}
The pre-registered parser (\S\ref{sec:protocol}) rejects any row whose token count is not $n$,
because a row of the wrong length cannot be aligned to columns. Most such rows miss by one
symbol, and in that case many of their cells are nonetheless determined. Let a row have its
\texttt{-} at the position of its own event, and let its length differ from $n$ by $k$. Reading
the row anchored at its left end assigns token $j$ to column $j$; reading it anchored at its
right end assigns token $j+k$ to column $j$. Wherever the two readings agree, the cell does not
depend on where the spurious or missing symbol lies, and we accept it; elsewhere we abstain.
Within a run of identical symbols the two readings always agree, which is why the rule recovers
a large share of long rows. A second pass fills any still-missing cell $(i,j)$ from the model's
own answer at $(j,i)$, converted to its converse. Neither rule ever overwrites a cell the strict
parser filled, and on cells both parsers fill they agree exactly (rate $1.000$ for all five
models), so the recovery parser yields a superset of the strict one.

\begin{table}[!ht]
\centering
\small
\setlength{\tabcolsep}{3.0pt}
\begin{tabular}{l cc cc cc}
\toprule
& \multicolumn{2}{c}{Coverage} & \multicolumn{2}{c}{Accuracy} & \multicolumn{2}{c}{Macro-F1} \\
\cmidrule(lr){2-3}\cmidrule(lr){4-5}\cmidrule(lr){6-7}
Model & strict & rec. & strict & rec. & strict & rec. \\
\midrule
\Ds & 0.600 & 0.836 & 0.452 & 0.630 & 0.270 & 0.301 \\
\Gp & 0.481 & 0.870 & 0.350 & 0.580 & 0.215 & 0.243 \\
\Gm & 0.577 & 0.979 & 0.328 & 0.535 & 0.128 & 0.150 \\
\Qw & 0.345 & 0.636 & 0.202 & 0.361 & 0.119 & 0.155 \\
\Ms & 0.024 & 0.116 & 0.001 & 0.039 & 0.003 & 0.014 \\
\bottomrule
\end{tabular}
\caption{Strict (pre-registered) versus recovery parsing of the same stored outputs. Accuracy
and macro-F1 are over all 101,880 pairs with missing cells counted as wrong. \emph{Always before}
reaches 0.607 accuracy.}
\label{tab:recovery}
\end{table}

Table~\ref{tab:recovery} gives the comparison. Accuracy on the cells that recovery adds but
strict parsing lacks is 0.767 (\Ds), 0.610 (\Gp), 0.516 (\Gm) and 0.555 (\Qw), against 0.753,
0.728, 0.569 and 0.585 on strictly parsed cells, so for the strongest model the recovered cells
are the more accurate of the two sets. Cells filled by the converse pass are weaker (0.676,
0.516, 0.462, 0.501). Comparing \Ds\ under recovery with \emph{always before} over a passage
bootstrap gives a difference of $+0.023$ with interval $[-0.041, +0.086]$ ($p=0.454$): parity,
not an advantage.

\paragraph{Common-subset scores.}
Answered-cell scores condition on a model-specific selection. Restricting all four models to the
22,665 cells (22.2\% of pairs) that every one of them answers removes that confound.
\Ds\ then reaches 0.805 accuracy and 0.438 macro-F1, \Gp\ 0.786 and 0.403, \Qw\ 0.642 and 0.268,
\Gm\ 0.614 and 0.230, against 0.615 and 0.059 for \emph{always before} on the same cells. The
ranking matches the primary metric, and the gap to the baseline is wider than on all pairs.

\paragraph{Precision and recall.}
Since a missing cell is a false negative and never a false positive, the two sides of macro-F1
behave differently: macro-precision is 0.387 (\Ds), 0.373 (\Gp), 0.301 (\Qw) and 0.291 (\Gm),
while macro-recall is 0.214, 0.156, 0.079 and 0.104.

\section{Controls for the Repair and Composition Analyses}
\label{app:controls}

\paragraph{A trivial fill beats the solver.}
The consistent repair fills missing cells with labels that make the network satisfiable, and the
gain in accuracy could come from the logic or from the skewed label prior. The control
substitutes a rule with no logic in it: fill every missing cell with \rel{<} and change nothing
else. It wins on every model and both metrics (Table~\ref{tab:repaircontrol}). Satisfiability
therefore buys less than majority guessing on this corpus.

\begin{figure}[!ht]
\centering
\includegraphics[width=\columnwidth]{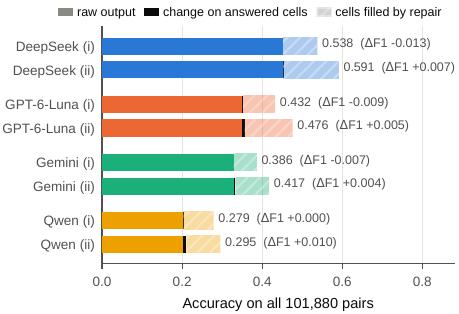}
\caption{Accuracy after the best-found consistent repair, decomposed into raw accuracy, the change on cells the model answered, and correct labels in cells the model left missing. (i): upper triangle as evidence; (ii): both triangles. $\Delta$F1 is the change in macro-F1.}
\label{fig:repair}
\end{figure}

\begin{table}[!ht]
\centering
\small
\setlength{\tabcolsep}{2.1pt}
\begin{tabular}{l ccc ccc}
\toprule
& \multicolumn{3}{c}{Accuracy} & \multicolumn{3}{c}{Macro-F1} \\
\cmidrule(lr){2-4}\cmidrule(lr){5-7}
Model & raw & rep.\,(ii) & fill \rel{<} & raw & rep.\,(ii) & fill \rel{<} \\
\midrule
\Ds & 0.452 & 0.591 & \textbf{0.702} & 0.270 & 0.278 & \textbf{0.284} \\
\Gp & 0.350 & 0.476 & \textbf{0.659} & 0.215 & 0.220 & \textbf{0.232} \\
\Gm & 0.328 & 0.417 & \textbf{0.582} & 0.128 & 0.132 & \textbf{0.142} \\
\Qw & 0.202 & 0.295 & \textbf{0.594} & 0.119 & 0.129 & \textbf{0.147} \\
\bottomrule
\end{tabular}
\caption{The best-found consistent repair (variant ii) against a control that fills every missing
cell with \rel{<}. The control is better everywhere.}
\label{tab:repaircontrol}
\end{table}

\paragraph{What the repair does to answered cells.}
Among answered cells the solver changes, the share moving from wrong to right is 0.42 (\Ds),
0.55 (\Gp), 0.39 (\Qw) and 0.22 (\Gm) in variant ii, and 0.26, 0.37, 0.27 and 0.21 in variant i;
for \Gm\ in variant i the solver breaks 371 correct cells while fixing 351.

\paragraph{Composition availability on correct cells.}
A composition-available error requires an intermediate event whose two relations the model
answered correctly and whose composition is single-valued and equals gold. If such configurations
were rare in general, the low rate among errors (\S\ref{sec:errors}) would say nothing about
composition. They are not rare: on cells the model answers correctly the same configuration is
available for 78.1\% (\Ds), 73.2\% (\Gp), 70.8\% (\Gm) and 66.2\% (\Qw) of cells, against 4.1\%,
2.5\%, 1.3\% and 1.9\% on errors. The contrast, a factor of 19 to 55, is the evidence that
failures are local to the pair rather than failures to combine available premises.

\paragraph{The cue interaction.}
\S\ref{sec:order} reports a GEE in which the gold relation is omitted. The reason is exact
collinearity: scored pairs are ordered $i<j$ in list order, so on before/after pairs gold \rel{>}
and ``the list presents the later event first'' are the same indicator. The model fitted is
$\textsc{correct} \sim \textsc{list} \times \textsc{narr} + \textsc{sentdist}$ with a logistic
link, independence working correlation and clustering by passage, on answered before/after pairs.

\section{Error Severity and Passage Size}
\label{app:severity}

\begin{figure}[!ht]
\centering
\includegraphics[width=\columnwidth]{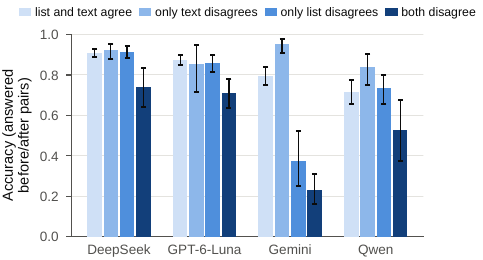}
\caption{Accuracy on answered before/after pairs, split by whether event list order and narration order agree with chronology (95\% passage-bootstrap CIs). Cell sizes for \Ds: 34,447, 1,902, 2,068 and 8,412.}
\label{fig:cue}
\end{figure}

\paragraph{Severity of answered errors.}
Each Allen relation fixes the order ($<$, $=$, $>$) of four endpoint pairs: start-start,
start-end, end-start and end-end (Table~\ref{tab:allen}). The severity of an answered error is
how many of those four comparisons differ between the predicted and the gold relation, so
\rel{m} for \rel{<} has severity 1 and \rel{>} for \rel{<} has severity 4. Severity 1 accounts
for 23.8\% (\Gp) to 33.4\% (\Gm) of answered errors, severity 2 is the most frequent class for
\Ds\ (43.5\%) and \Gp\ (42.0\%), and severity 4 accounts for 19.4\% (\Ds) to 30.0\% (\Qw).
Near misses dominate only for particular gold relations: 83.6\% of \Ds's errors on gold \rel{o}
and 65.5\% on gold \rel{m} have severity 1, against 4.0\% on gold \rel{>}. Figure~\ref{fig:factors}
and Figure~\ref{fig:confusion} give the full breakdowns.

\paragraph{Passage size and output length.}
Passage-level macro-F1 correlates negatively with the number of events (Spearman $\rho$ from
$-0.57$ for \Ds\ to $-0.68$ for \Qw) and with word count ($-0.41$ to $-0.52$); the regression
reported in \S\ref{sec:length} separates the two. Macro-F1 for \Gp\ falls from 0.463 on the
smallest passages to 0.033 above 40 events. Within an output, accuracy declines from the first
to the last decile of generated rows for \Gp\ (0.410 to 0.291) and \Qw\ (0.244 to 0.164), while
\Ds\ shows no such decline (Figure~\ref{fig:factors}c).

\section{Relative Time in the Corpus}
\label{app:relative}

The passages locate events relative to other events rather than on a calendar. We quantify this and explain why it makes the task hard.

\paragraph{Measurement.}
We profiled the passages with regular expressions (Table~\ref{tab:relative}); the counts are heuristic, not annotations. \emph{Absolute anchors} are expressions that fix a position on the calendar: a year or a day of a month. \emph{Partial anchors} name a cyclic unit without fixing its position (weekdays, month names without a year, clock times). \emph{Offsets} state a distance from another time (``twelve days later'', ``three months ago''). \emph{Deictic phrases} locate a time relative to a reference point (``the following week''). \emph{Connectives} include \emph{before}, \emph{after}, \emph{when}, \emph{while}, \emph{until}, \emph{since} and \emph{by then}. \emph{Past perfect} (\emph{had} plus participle) marks an event placed before the current narrative reference time.

\begin{table}[!ht]
\centering
\small
\setlength{\tabcolsep}{4pt}
\begin{tabular}{l r r r}
\toprule
Type & Mentions & \makecell{Per 100\\words} & \makecell{Passages\\with $\geq$1} \\
\midrule
Absolute anchor & 38 & 0.04 & 22 (6.1\%) \\
Partial anchor & 919 & 0.89 & 197 (54.7\%) \\
Offset & 494 & 0.48 & 246 (68.3\%) \\
Deictic phrase & 598 & 0.58 & 211 (58.6\%) \\
Connective & 2,488 & 2.42 & 358 (99.4\%) \\
Past perfect & 1,909 & 1.86 & 300 (83.3\%) \\
\bottomrule
\end{tabular}
\caption{Temporal expressions in the 360 passages (102,746 words), counted with regular expressions. Categories are defined in the text.}
\label{tab:relative}
\end{table}

Absolute anchors are rare: 35 of the 38 are bare years, three name a day of a month, and 338 passages (93.9\%) have none. Partial anchors are more common but do not fix positions either, and 157 passages (43.6\%) contain no calendar expression of any kind. Only 4.1\% of the 7,627 events occur in a sentence with an absolute anchor and 24.0\% in a sentence with any calendar expression, whereas 58.0\% of events occur in a sentence with at least one relative cue. Relations must therefore be inferred from offsets, connectives, tense and narrative structure.

\paragraph{Why relative time is hard.}
Seven properties of relative time make the task hard, and our results bear on each.
\begin{enumerate}[leftmargin=*,itemsep=1pt,topsep=2pt]
\item \textbf{No shared axis.} With absolute anchors, each event can be placed independently and any pair can be ordered by comparing two positions, which is why anchoring events to times scales linearly with the number of events \citep{reimers-etal-2016-temporal} and why timeline systems normalise explicit time expressions first \citep{do-etal-2012-joint}. With relative expressions, an event's position is defined only through other events, so the relation of a distant pair must be assembled from a chain of intermediate relations. In our gold data 88.7\% of pairs are derivable by one composition step, yet composition-available errors are rare (\S\ref{sec:errors}): models seldom have the correct premises in the first place, and one wrong link corrupts every relation derived through it.
\item \textbf{Explicit times are scarce even in standard corpora.} Only about 20\% of event mentions in TimeBank-Dense are linked to a time expression \citep{wen-ji-2021-utilizing}, and pairwise annotation schemes leave 58\% of TimeBank events with imprecise temporal information \citep{reimers-etal-2016-temporal}. Our passages are more extreme, with absolute anchors near 4\% of events.
\item \textbf{Models lean on explicit anchors.} When absolute dates in biographies are rewritten as relative or event-anchored expressions, ordering accuracy falls for humans and for most models, and masking a single year can disrupt an otherwise correct order \citep{sun-etal-2025-time}. Benchmarks that combine numerical and relational reasoning report that models fall back on heuristic shortcuts rather than grounded deduction \citep{qian-etal-2026-beyond}.
\item \textbf{Relative expressions must be resolved.} An offset such as ``two weeks after'' is interpretable only once its reference event is found, and the past perfect is anaphoric to a reference time set elsewhere \citep{kehler-1994-temporal}. Temporal signals are also ambiguous: \emph{in} acts as a temporal signal only 13.3\% of the time in TimeBank \citep{derczynski2012corpusbasedstudytemporalsignals}. Partial anchors add their own ambiguity, since ``on Tuesday'' and ``that Friday'' cannot be ordered without knowing which week each refers to.
\item \textbf{The default is narration order.} When no explicit cue overrides it, readers interpret successive clauses as successive events \citep{kehler-1994-temporal}, and several resources treat narration order as a proxy for chronology \citep{yao-huang-2018-temporal,lin-etal-2021-conditional}. Models inherit this narrative bias \citep{fang-etal-2024-getting}. Flashbacks are signalled mainly by tense and connectives (past perfect in 83.3\% of passages), and our models fail most when narration and event list both present the later event first (\S\ref{sec:order}).
\item \textbf{Interval boundaries are implicit.} Allen relations depend on when events end as well as when they start, but narratives rarely state end points or durations, which readers infer from commonsense \citep{zhou-etal-2019-going}; such knowledge is under-reported in text \citep{zhou-etal-2020-temporal}. Relations defined by shared endpoints (\rel{m}, \rel{s}, \rel{f}, \rel{o}) are thus rarely stated outright, matching their low F1 (\S\ref{sec:relations}).
\item \textbf{Offsets require arithmetic along chains.} Placing events via ``twelve days later'' and ``two weeks after'' means adding durations along a chain, a known weakness of LLMs \citep{islakoglu-kalo-2025-chronosense,chu-etal-2024-timebench,wang-zhao-2024-tram}.
\end{enumerate}

\paragraph{Do the available anchors help?}
If calendar expressions supplied the missing axis, pairs whose events both occur in sentences with one should be easier. They are not: on answered cells, \Ds\ reaches 0.720 accuracy on such pairs and 0.748 on pairs where neither event has one, and \Gp\ 0.662 versus 0.728. As these are almost all partial anchors, they rarely fix positions; the comparison is observational, as anchors cluster in some passage types.

\end{document}